\documentclass[conference]{IEEEtran}
\IEEEoverridecommandlockouts
\usepackage{cite}
\usepackage{amsmath,amssymb,amsfonts}
\usepackage{algorithmic}
\usepackage{graphicx}
\usepackage{textcomp}
\usepackage{xcolor}
\usepackage{subfigure}
\usepackage{amsthm,amsmath,amssymb}
\usepackage{mathrsfs}
\usepackage{bm}
\usepackage{float}
\usepackage{CJKutf8}
\usepackage{makecell}
\usepackage{multirow}
\usepackage{booktabs}

\graphicspath{{Fig/}}
\def\BibTeX{{\rm B\kern-.05em{\sc i\kern-.025em b}\kern-.08em
    T\kern-.1667em\lower.7ex\hbox{E}\kern-.125emX}}
\begin{document}

\title{A Simple Baseline for Adversarial Domain Adaptation-based Unsupervised Flood Forecasting
\thanks{This work was partially funded by Natural Science Foundation of Jiangsu Province under grant No. BK20191298, Science and Technology on Underwater Vehicle Technology Laboratory, Research Fund from Science and Technology on Underwater Vehicle Technology Laboratory(2021JCJQ-SYSJJ-LB06905), Water Science and Technology Project of Jiangsu Province under grant No. 2021072.}
}

\author{
\IEEEauthorblockN{Delong Chen, Ruizhi Zhou, Yanling Pan, and Fan Liu}
\IEEEauthorblockA{College of Computer and Information, Hohai University, Nanjing, China\\
chendelong@hhu.edu.cn}}

\maketitle

\begin{abstract}
    Flood disasters cause enormous social and economic losses. However, both traditional physical models and learning-based flood forecasting models require massive historical flood data to train the model parameters. When come to some new site that does not have sufficient historical data, the model performance will drop dramatically due to overfitting. This technical report presents a \textbf{Flood} \textbf{D}omain \textbf{A}daptation \textbf{N}etwork (FloodDAN), a baseline of applying \textbf{U}nsupervised \textbf{D}omain \textbf{A}daptation (UDA) to the flood forecasting problem. Specifically, training of FloodDAN includes two stages: in the first stage, we train a rainfall encoder and a prediction head to learn general transferable hydrological knowledge on large-scale source domain data; in the second stage, we transfer the knowledge in the pretrained encoder into the rainfall encoder of target domain through adversarial domain alignment. During inference, we utilize the target domain rainfall encoder trained in the second stage and the prediction head trained in the first stage to get flood forecasting predictions. Experimental results on Tunxi and Changhua flood dataset show that FloodDAN can perform flood forecasting effectively with zero target domain supervision. The performance of the FloodDAN is on par with supervised models that uses 450-500 hours of supervision.
\end{abstract}


\section{Introduction}
    \label{sec:introduction}
    Flood is one of the most severe issues threatening human security and economic development. Timely and accurate flood forecasting can make a great contribution to prevention countermeasures against floods, saving more lives, and reducing more losses. Nowadays, there are mainly two kinds of methods to predict floods: physical model \cite{DBLP:journals/envsoft/CostabileM15,Collischonn2005ForecastingRU, Salas2018TowardsRC} and data-driven learning-based model \cite{Cheng2015DailyRR,SHOAIB2016211}. The physical model relies on complex hydrological parameters and intensive computation to predict floods. However, it is heavily based on hydrology knowledge, which limits its generalization ability and scalability. Data-driven models is primarily based on the learning of the relationship between the input data and observed targets, without explicitly involving of any human-defined physical parameters. Compared to the physical model, the data-driven model can achieve higher generalization in a diversity of rivers \cite{DBLP:journals/corr/abs-1908-02781}. In recent years, with the development of machine learning, using learning-based models \cite{Sulaiman18Heavy,DBLP:journals/eswa/GuoZQZL11,SHU200831,Nourani2014ApplicationsOH} has gained high popularity and shows better performance than traditional approaches \cite{DBLP:journals/corr/abs-1908-02781}.

However, data-driven learning-based models usually require massive training data. Globally, there are many rivers with little historical hydrologic data, especially in developing countries and rivers on small scale. Hence, predicting floods with limited hydrologic data source is rather necessary for those areas. There are some related works \cite{DBLP:journals/corr/abs-2009-14379} that have made efforts to solve this problem. By far, most solutions focus on adding spatiotemporal sequence data and adjusting network architectures to achieve better performance. But this kind of approach still requires a certain amount of labels. Therefore, it is of great significance to predict floods with zero supervision, which is a extremely challenging task to tackle.

In this technical report, we propose a \textbf{Flood} \textbf{D}omain \textbf{A}daptation \textbf{N}etwork (FloodDAN) approach for unsupervised flood forecasting, which integrates large-scale pretraining and adversarial domain adaptation to generate a model for flood forecasting. We first pretrain the source model with a large-scale dataset, then perform adversarial domain adaptation between two datasets. Finally, we adopt the target encoder generated in stage 2 and the source prediction head generated in stage 1 to build the final model. From the experimental results, FloodDAN achieves flood forecasting with only rainfall data. Due to the measurement of runoff has higher technical requirements, while measuring rainfall is simple. Our proposed approach has great value in practical application.

The main contributions of the technical report can be concluded below.

\begin{itemize}
    \item We show that flood forecasting can be done in an unsupervised manner. In our proposed FloodDAN, we first utilize  large-scale pretraining dataset to learn hydrological knowledge, then transfer the knowledge to target domain by adversarial learning
    
    \item We conduct experiments on Tunxi and Changhua flood dataset.  Experimental results show that our proposed FloodDAN can perform flood forecasting effectively with zero target domain supervision, achieving performance that is on par with supervised models that uses 450-500 hours of supervision.
    
\end{itemize}

The rest of the technical report is organized as follows. Section 2 reviews related work. Section 3 introduces our model structure and presents details of our model. Section 4 presents experimental results. Finally, Section 5 concludes this technical report.

\section{Related Works}
    \label{sec:relatedworks}
    In this section, we review several researches related with our work, mainly including learning-based approaches and domain adaptation based approaches.
    
\subsection{Learning-based Flood Forecasting}

    In the past decade, with the development of machine learning and deep learning, learning-based methods have been deeply applied in natural disaster prediction. For example, Wavelet Graph Neural Network (WGNN) \cite{9515293} is an attempt to predict tsunamis. For flood forecasting tasks, compared to conventional physical models, learning-based methods can formulate nonlinear flood forecasting relationships without reliance on the hydrological knowledge of rivers. This kind of method can achieve high performance with less complexity based on historical datasets \cite{DBLP:journals/corr/abs-1908-02781}.
    
    Learning-based approaches to predict flood include SAE-BP \cite{7966716}, Artificial Neural Networks (ANNs) \cite{Sulaiman18Heavy}, Wavelet Neural Network (WNN) \cite{Nourani2014ApplicationsOH}, Adaptive Neuro-Fuzzy Inference System (ANFIS) \cite{SHU200831,Dineva14Fuzzy}, Support Vector Machine (SVM) \cite{DBLP:journals/eswa/GuoZQZL11}. Those methods can be briefly classified into two types, single ML-based approaches and hybrid ML-based approaches. The first group uses individual ML approaches while the second group adopts soft computing techniques, statistical methods, and physical models to improve their performance.
    
    Sulaiman et al. \cite{Sulaiman18Heavy} applied ANN to formulate the relationship between precipitation and flood. They based on local precipitation data from 1965 to 2015 to train the ANN model and showed the reliability of the model in forecasting risky flood events. However, ANNs is relatively less accurate and sometimes falls into overfitting. Guo et al. \cite{DBLP:journals/eswa/GuoZQZL11} used SVM to forecast monthly streamflow and their experiments showed that their improved SVM model can achieve higher prediction accuracy and better generalization ability. Shu et al. \cite{SHU200831} proposed a methodology of using ANFIS for flood quantile estimation. This hybrid machine learning-based method is easy to implement and has strong generalization ability.
    
    Although those approaches above achieve great performance, they all have a drawback that they need massive data to build a reliable model and their performance falls short if the historical hydrological data is limited. This problem has been existed for a long time and remains a challenging task. Therefore, we consider promoting an advanced ML-based approach that integrates domain adaptation to solve this problem.

\subsection{Deep Domain Adaptation}
   
    As one of the transfer learning methods, the domain adaptation method sheds light on the scenario that the distributions of source datasets and target datasets are different \cite{pan2009survey}. For example, Long et al. \cite{long2015learning} proposed a Deep Adaptation Network (DAN) that matches the changes in the cross-domain marginal distribution by adding multiple adaptation layers and multiple kernels. They also \cite{long2016unsupervised} proposed a new method for domain adaptation in deep networks. This method assumes that there exists a residual function between the source classifier and the target classifier. It can learn adaptive classifiers and transferable features from the source domain and the target domain. 
    
    There are a series of researches that apply domain adaptation in many scenarios. Chen et al. \cite{DBLP:journals/corr/ChenLCHFS17} proposed an adversarial training procedure to minimize the differences between domains, that performed well on a series of datasets and achieved great improvement after adaption. Ghoshal et al. \cite{DBLP:journals/corr/abs-2005-00791} applied Domain-Adversarial Neural Networks (DANNs) \cite{DBLP:conf/icml/GaninL15,DBLP:journals/jmlr/GaninUAGLLML16} in cross-domain sentiment analysis task. They introduced a new framework that utilized the ConceptNet knowledge graph to enhance the sentiment analysis among the Books, Electronics, DVD, and Kitchen domains. The proposed framework outperformed state-of-the-art approaches in most scores. Many other applications can be seen in literature \cite{DBLP:journals/ijon/WangD18,DBLP:journals/corr/abs-2010-03978,DBLP:conf/coling/RamponiP20,DBLP:journals/corr/abs-2102-09508}.
    
    However, by far, few researchers have applied domain adaption methods in flood forecasting. Therefore, this paper proposes a novel, fusion domain adaptive deep learning flood forecasting framework. The framework narrows the distribution difference between the source domain and the target domain through domain adaptation. Hence, our FloodDAN can utilize the large-scale hydrological dataset to predict floods through unsupervised learning, whereby the generalization ability of our approach can be improved as well.

\section{Approach}
    \label{sec:approach}
    \subsection{Problem Formulation}
    Suppose we have a dataset $\mathcal{D}_{\text{source}}=\{({X}_i, {Y}_{i}^{\text{history}}, {Y}_i)\}_{i=1}^N$ with $N$ samples, where ${X}_i$ is the $i$th input rainfall data, ${Y}_{i}^{\text{history}}$ is the $i$th input historical runoff data and ${Y}_i$ is the ($T+t$) th runoff data. $t$ denotes the forecast period, which means the duration between input data and output data. ${X}_i$ is a $d_{\text{source}}$-dimensional vector. $d_{\text{source}}$ denotes the number of source hydropower stations. In stage 1, our goal is to pretraining a rainfall encoder $E_{\text{source}}$ and a prediction head $h_{\text{source}}$ to predict ${Y}_i$ from ${X}_i$ and ${Y}_{i}^{\text{history}}$ accurately. 
    
    In addition, we have a target domain dataset $\mathcal{D}_{\text{target}}=\{({X}_j,{Y}_{j}^{\text{history}}\}_{j=1}^M$ with $M$ samples, where ${X}_j$ is the $j$th input rainfall data and ${Y}_{j}^{\text{history}}$ is the $j$th input historical runoff data. Note that the historical runoff data is only visible during model inference. ${X}_j$ is a $d_{\text{target}}$-dimensional vector, where $d_{\text{target}}$ denotes the number of target hydropower stations. In stage 2, our goal is to learn a encoder $E_{\text{target}}$ which can extract the features of $X_{j}$.

\subsection{FloodDAN Learning Procedure}
    \begin{figure*}[t]
    \centering
        \includegraphics[width=0.8\linewidth]{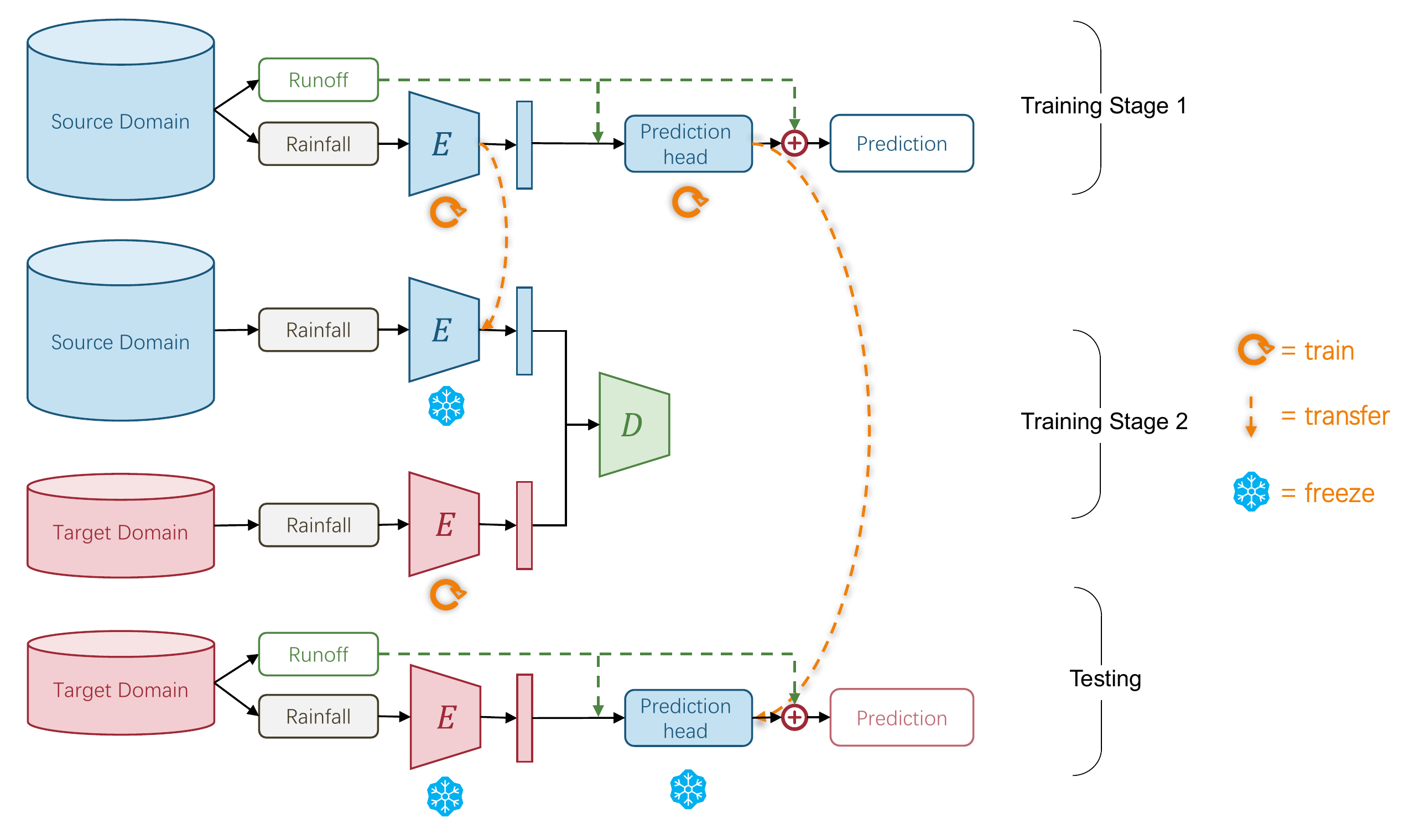}
        \caption{The pipeline of our FloodDAN. In stage 1, we pretrain a source model with the source dataset in the supervised manner. In stage 2, we train a target encoder through adversarial domain adaptation. Last, we combine the target encoder and the source prediction head to predict the runoff of the target domain.}
        \label{fig:pipeline}
    \end{figure*}
    
    An overview of our approach is shown in Fig. \ref{fig:pipeline}. First of all, we pretraining a source encoder and a source prediction head with a large-scale source dataset. Then, we use adversarial learning to train a target encoder which used to extract the features of the target dataset. Last, we splice the target encoder and the source prediction head as the final model and test its performance.

\subsubsection{Training Stage 1 -- Large-scale Pretraining}
    For datasets without labels, there is no supervised way to learn the relationship between inputs and outputs. However, part of the relationship between rainfall and runoff is transferable between different hydrological datasets. For example, increased rainfall usually leads to greater runoff; Under the condition of equal rainfall, large historical runoff will lead to larger future runoff. So in the first stage, we pretraining a model with a large-scale dataset to learn the function, which is able to map the rainfall to the runoff.

    We use mean squared error (MSE) as the loss function, which is defined as:
    \begin{equation}
        \mathcal{L} = \sum_{i=1}^N ||{Y}_{i}-\hat{Y_{i}}||^2
    \end{equation}
    
    After pretraining, we get a source encoder which can effectively extract the features of rainfall, and a source prediction head which can map the features and historical runoff to the future runoff. 
    
\subsubsection{Training Stage 2 -- Adversarial Domain Adaptation}
    Due to the extensive training samples, the source model has learned the relationship between rainfall and runoff. However, although it can extract the features related to the runoff from the rainfall, it is not accurate to transfer it to another dataset for flood forecasting. Because different datasets are in different domains. For example, source domain has little rainfall, while target domain has a lot. If we use the source model to predict the runoff of target domain, the predicted runoff will be much smaller than the ground truth based on the knowledge learned from the source domain.  
    
    Hence, we have to train a target encoder. In stage 2, we freeze the source encoder and use a domain discriminator $D$ to score both target feature $F_{target}$ and source feature $F_{source}$. The purpose of this is to make the output of the target encoder match that of the source encoder as much as possible. It means that the target encoder can learn the feature of the target rainfall without accepting any labels. And through adversarial learning, source features and target features will eventually be distributed in the same feature space. That is to say, domain adaptation is implemented.
    
    In our FloodDAN, the target encoder acts as a generator. The generator generates features by $\hat{{Y}_{i}}=E({X}_{i})$. Its loss function is the adversarial loss, which is shown in Eq.\ref{eq:G_loss}.
    
    \begin{equation}
        \mathcal{L}_{G} = - \sum_{i=1}^N {D(\hat{{Y}_{i}})}
        \label{eq:G_loss}
    \end{equation}
    
   Eq.3 shows the loss function of the discriminator. The third term is the gradient penalty term of Wasserstein GAN~\cite{Gulrajani2017NIPS_improved}, where $\tilde{\bm{Y}}$ is obtained by random linear interpolation between $\hat{\bm{Y}}$ and $\bm{Y}$ and $w_{GP}$ is the weight for the gradient penalty term.
    
    \begin{equation}
        \mathcal{L}_{D} = \sum_{i=1}^N D(\hat{{Y}}_i) - D({Y}_i) + w_{GP}\mathbb{E}_{\tilde{\bm{Y}}} ||\nabla_{\tilde{\bm{Y}}}D(\tilde{\bm{Y}})-1||^2_2
        \label{eq:D_loss}
    \end{equation}
    
\subsubsection{Inference -- Unsupervised Flood Forecasting}

After training, the target encoder can take advantage of prior knowledge learned in pretraining to extract target features. And through adversarial learning, target features and source features are in the same domain. Therefore, we can combine the target encoder and the source prediction head to predict the target runoff. 

In this case, our FloodDAN realize unsupervised learning for flood forecasting. It can be applied to forecast floods in a place that lacks historical hydrological data, and the results are not bad.

\subsection{Model Structure}
    
    \begin{figure}[t]
    \centering
        \includegraphics[width=0.48\textwidth]{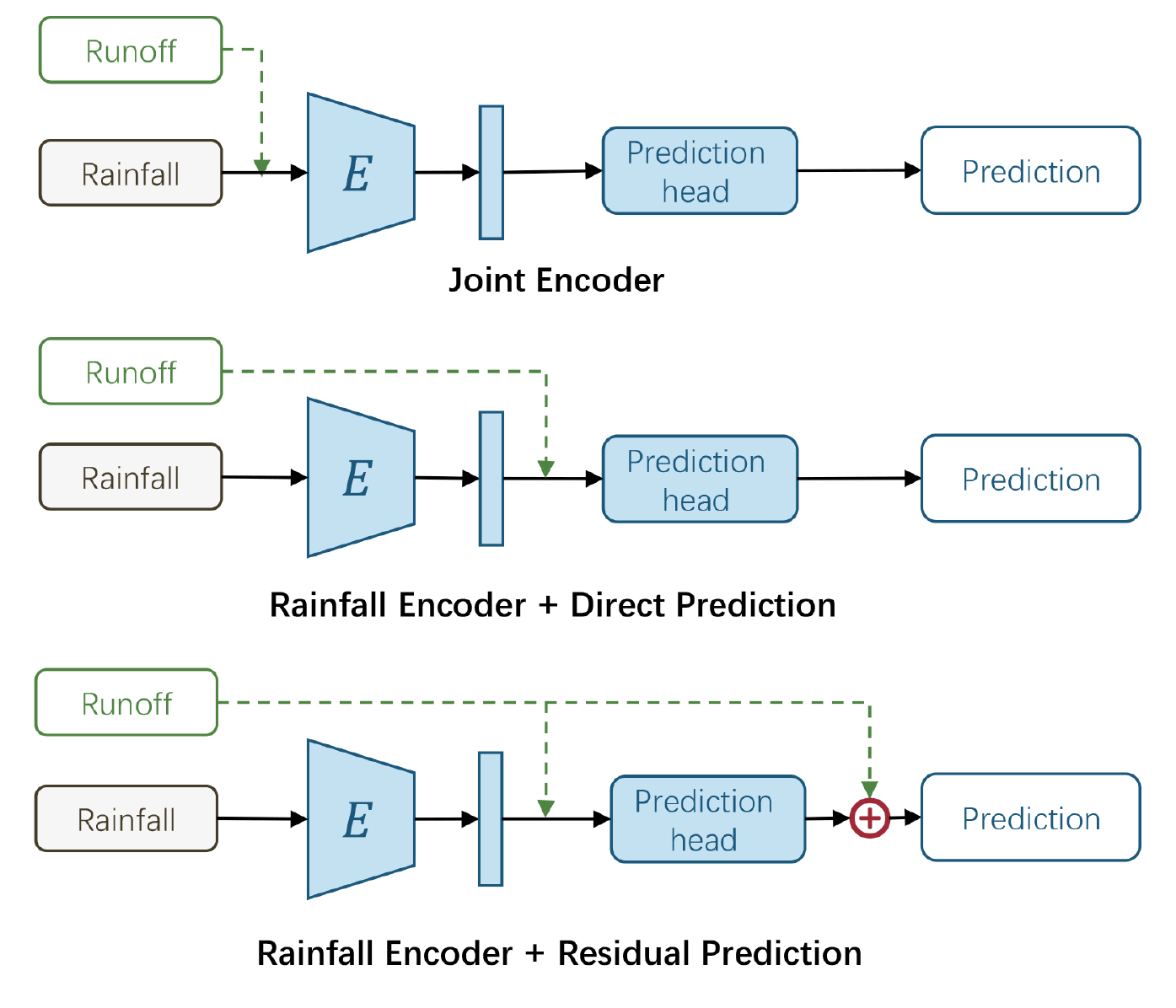}
        \caption{Different model structures for flood forecasting.}
        \label{fig:model}
    \end{figure}
    
    Future runoff is related to historical runoff and rainfall. So we decide to design an encoder to extract their features. Once we have these features, we need to map them to future runoff for prediction purposes. There are a prediction head should be added. Finally, we have to determine how the data flows through the model. We considered three options for this problem, which are shown in Fig. \ref{fig:model}
    
    The aim of this work is to achieve flood forecasting in areas where historical runoff data are not available. If both historical rainfall and runoff are fed into the encoder, the target encoder cannot be learned in stage 2 because of the lack of runoff data. Therefore, in order to achieve unsupervised learning, runoff should be input into the prediction head together with extracted rainfall features. To prevent network degradation, we add residual connection in prediction head.
    
    Since flood forecasting is a sequential task, we compare temporal convolutional network (TCN), Long short-term memory (LSTM), Gated Recurrent Unit (GRU) and other models to find the most suitable encoder structure. Both LSTM and GRU can solve the problem of gradient disappearance of traditional RNN. And they can control long-term memory. But TCN has more advantages. First, TCN introduces dilated convolution and residual connection. We can adjust the receptive field size through the number of layers, dilated factor and filter size, which enables us to control the memory size of the model according to the requirements of different fields. Second, it has some unique advantages over LSTM and GRU because it uses convolution structure \cite{9451544}. In addition, many time series models have adopted TCN and achieved good performance, such as conducting motion generation \cite{jcst2022self}. Considering comprehensively, we choose TCN as encoder. On the premise of ensuring the good performance of the model, we try to make the model simple. We employ three TCN layers. Each of them has 36 filters of size 2 and a dropout of 0.2 with subsequent $ReLU$ as the activation function.
    
    The purpose of prediction head is to learn a function which can map the historical runoff and features to the future runoff. Therefore, we employ three 1D-CNN as the prediction head: the first two layers with 36 filters of size 2 with subsequent ReLU as the activation function; and a third with 1 filters of size 3 for reducing vector dimension.

\section{Experiments}
    \label{sec:experiment}
    \subsection{Datasets}
    \label{subsec:datasets}
    
In our experiment, we use hydrological data of two watersheds, Tunxi watershed (Anhui, China) and Changhua watershed (Zhejiang, China). The two datasets include hourly rainfall in every hydrological station and hourly stream flows of the two watersheds.

\textbf{Tunxi}: This dataset covers the data of 11 rainfall stations and 1 flow station from June 27, 1981 to July 4, 2003, including 43435 hourly samples. The catchment of Tunxi watershed is 2696.76 $\text{km}^{2}$. We use it as the source dataset.

\begin{figure}[t]
\centering
    \includegraphics[width=0.35\textwidth]{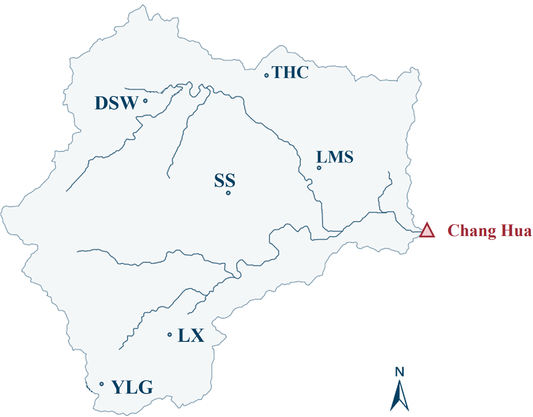}
    \caption{Map of Changhua dataset. There are 7 rainfall stations and 1 flow station.}
    \label{fig:changhua_map}
\end{figure}

\textbf{Changhua}: As shown in Fig.~\ref{fig:changhua_map}, this dataset covers the data of 7 rainfall stations and 1 flow station from 1998 to 2008, There are a total of 8159 hourly data. The catchment of Changhua watershed is 3444 $\text{km}^{2}$. We use it as the target dataset.

We set the train-test split ratio is 7:3. In order to avoid errors caused by data differences, all the experimental results in this paper are evaluated on the same testing set.

\subsection{Experiment Setup}
    
    We have implemented our approach by Pytorch. Forecast period is set to 6 hours and $w_{GP}$ is set to 10. We train the model for 100 epochs with a batch size of 64, a learning rate of 5e-4, and a weight decay of 8e-3. We reduce the learning rate by cosine learning rate scheduler. Both pretraining and adversarial domain adaptation use the same setup except for the optimizer. We use AdamW optimizer for pretraining but RMSprop for adversarial domain adaptation. What's more, the input data should be normalized to [0,1]. 
    
    \begin{table*}[h]
      \begin{center}
        \caption{ Experimental results under different settings}
            \begin{tabular}{ccccc}
              \toprule
              \textbf{Supervision} & \textbf{Model} & \textbf{Training data} & \textbf{MSE} ($\downarrow$) & \textbf{DC\%} ($\uparrow$) \\
              \midrule
              \multirow{8}*{Fully Supervised Machine Learning} & SVR & \multirow{8}*{5710 h} & 6463 & 68.00\\
               & Decision Tree &  &  9543 & 53.00 \\
               & Linear Regression &  &  6479 & 68.00  \\
               & k-NN &  &  7248 & 64.00  \\
               & Random Forest &  &  6900 & 66.00  \\
               & Gradient Boosting &  &  5980 & 70.00 \\
               & Bagging &  &  7519 & 63.00  \\
               & Joint Encoder &  &  6581 & 68.03  \\
              \midrule
              \multirow{2}*{Fully Supervised Deep Learning} & Rainfall Encoder & \multirow{2}*{5710 h} & 3846 & 81.13\\
               & Rainfall Encoder + Residual Prediction &  & 3733 & 81.82  \\
              \midrule
              \multirow{15}*{Few-shot Learning} & \multirow{15}*{Rainfall Encoder + Residual Prediction} & 114 h & 12038 & 41.96 \\
                & & 143 h & 9177 & 55.71 \\
                & & 171 h & 8964 & 56.80  \\
                & & 200 h & 8657 & 58.26  \\
                & & 228 h & 8211 & 60.27  \\
                & & 257 h & 9112 & 56.07  \\
                & & 285 h & 9114 & 56.06  \\
                & & 314 h & 8626 & 58.32  \\
                & & 342 h & 8359 & 59.67  \\
                & & 371 h & 8311 & 59.91  \\
                & & 399 h & 8122 & 60.79  \\
                & & 428 h & 8180 & 60.60  \\
                & & 456 h & 7733 & 62.63  \\
                & & 485 h & 7646 & 63.12  \\
                & & 513 h & 7135 & 65.61  \\
              \midrule
              \multirow{3}*{Unsupervised Learning} & Lower-bound & \multirow{3}*{0 h} & 11176 & 45.58\\
               & Rainfall Encoder + Direct Prediction (FloodDAN) &  & 7848 & 62.81  \\
               & Rainfall Encoder + Residual Prediction (FloodDAN) &  & 7705 & 63.49  \\
              \bottomrule
            \end{tabular}
      \end{center}
    \end{table*}
    
\subsection{Evaluation Metrics}

    We use MSE and deterministic coefficient (DC) as the evaluation metrics, whose definitions are respectively showed in Eq.\ref{eq:MSE} and Eq.\ref{eq:DC},
    
    \begin{equation}
        \textit{MSE} = \frac{1}{n} \sum_{i=1}^{t}[\hat{y_{i}} - y_{i}]^{2}
        \label{eq:MSE}
    \end{equation}
    
    \begin{equation}
        \textit{DC} = 1 - \frac{\sum_{i=1}^{t}[\hat{y_{i}} - y_{i}]^{2}}{\sum_{i=1}^{t}[y_{i} - \overline{y}]^{2}}
        \label{eq:DC}
    \end{equation}
    , where $t$ denotes the number of sample; $\hat{y_{i}}$ denotes the prediction of the $i$th sample; $y_{i}$ denotes the label of the $i$th sample; $\overline{y}$ denotes the average of the labels of all samples.
    
    MSE is a common evaluation metric for predictive tasks, it said the gap between the predicted results with the actual results. The lower it is, the closer the predicted results are to the ground truths. DC is often used to characterize the quality of a data fitting model. It measures how well regression models fit data. We know from Eq.\ref{eq:DC} that DC is a number less than 1. The closer it is to 1, the stronger the explanatory ability of the input to the output of the model is, and the better the model fits the data.
    
\subsection{Comparisons and Results}

    To validate the effeteness of our FloodDAN, we compare the performances of following approaches:
    
    \begin{itemize}
    \item \textbf{Fully supervised learning.} To find the best model structure for flood forecasting, we use the entire training set for supervised training and compare the results of evaluation. It also sets a baseline for the model.
    
    \item \textbf{Few-shot learning.} There is no baseline available to compare our approach because our approach is the first to use unsupervised learning for flood forecasting. Hence, we reduce the scale of the training set and train a model through supervised learning to set a baseline.
    
    \item \textbf{Unsupervised learning.} Using the input historical runoff as the model output, we calculate a lower-bound of unsupervised learning approach. And we use residual prediction model and direct prediction model respectively to implement our method for futher comparison. 
    \end{itemize}
    
    \begin{figure}[h]
        \centering
        \includegraphics[width=0.5\textwidth]{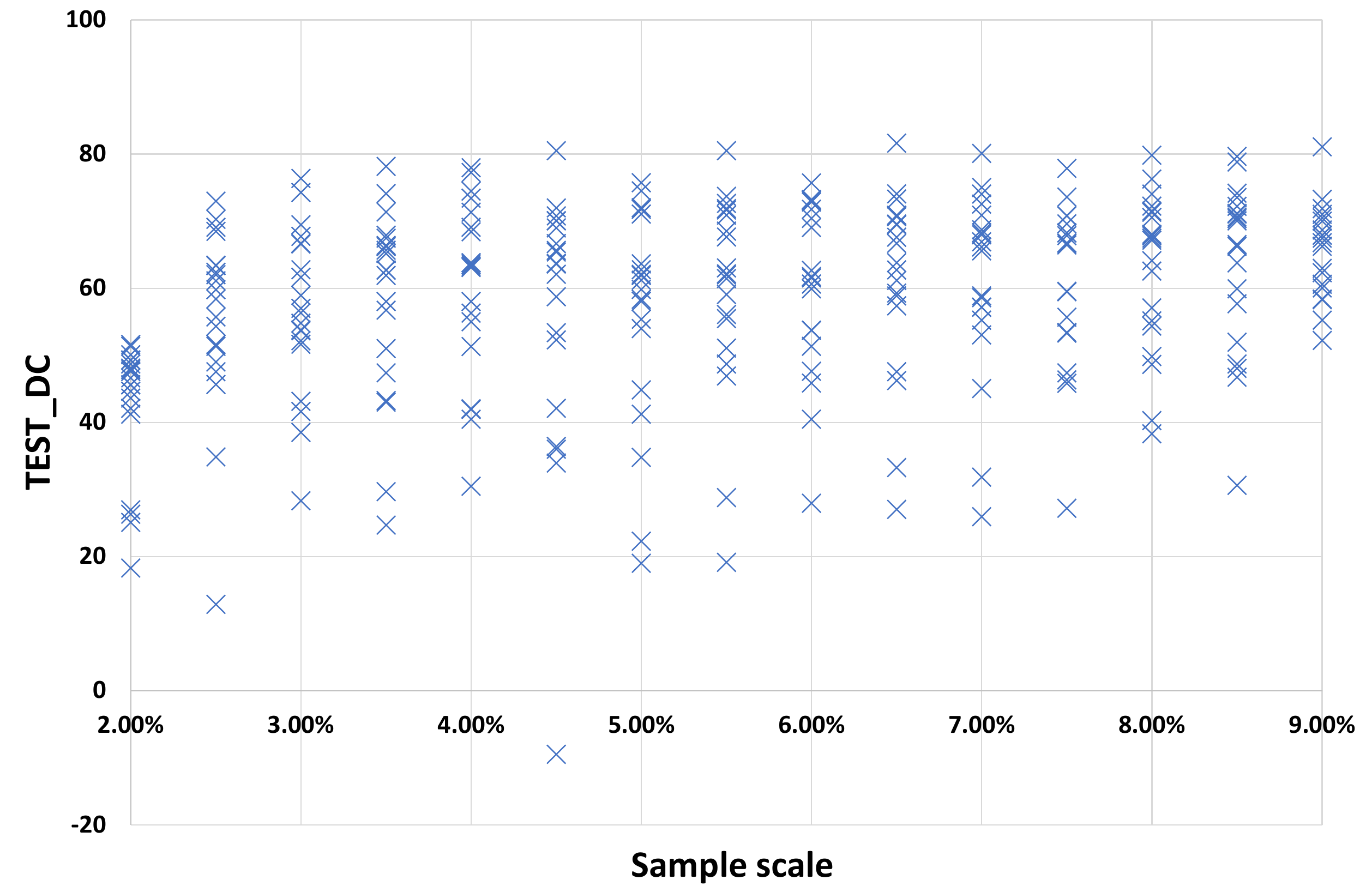}
        \caption{Performances of our models in few-shot supervised manner.}
        \label{fig:sample_scale}
    \end{figure}
    
    \begin{figure*}
    \centering
        \includegraphics[width=1\textwidth]{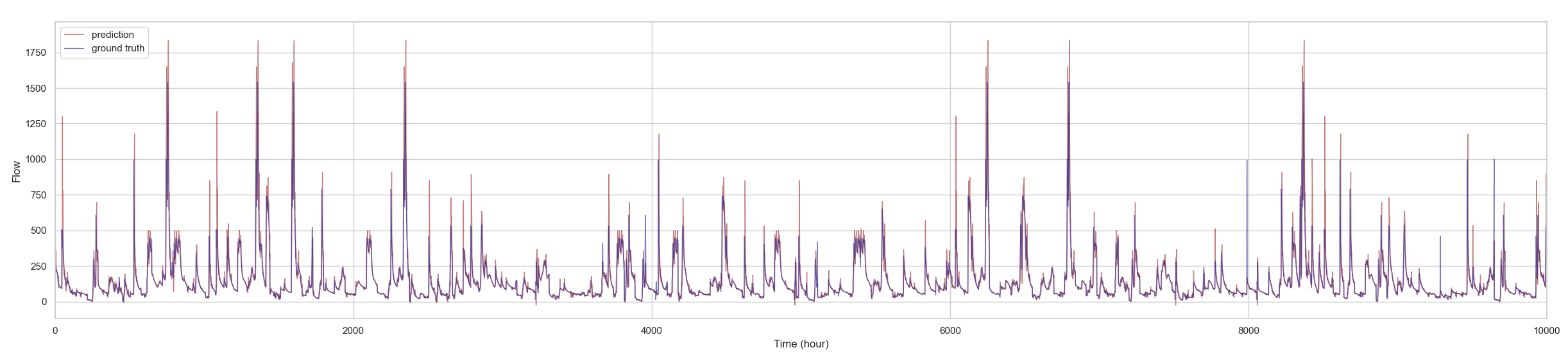}
        \caption{Test results of the model obtained by splicing changhua encoder and Tunxi prediction head on Changhua dataset.}
        \label{fig:Changhua_result}
    \end{figure*}
    
    \begin{figure}[h]
        \centering
        \includegraphics[width=0.35\textwidth]{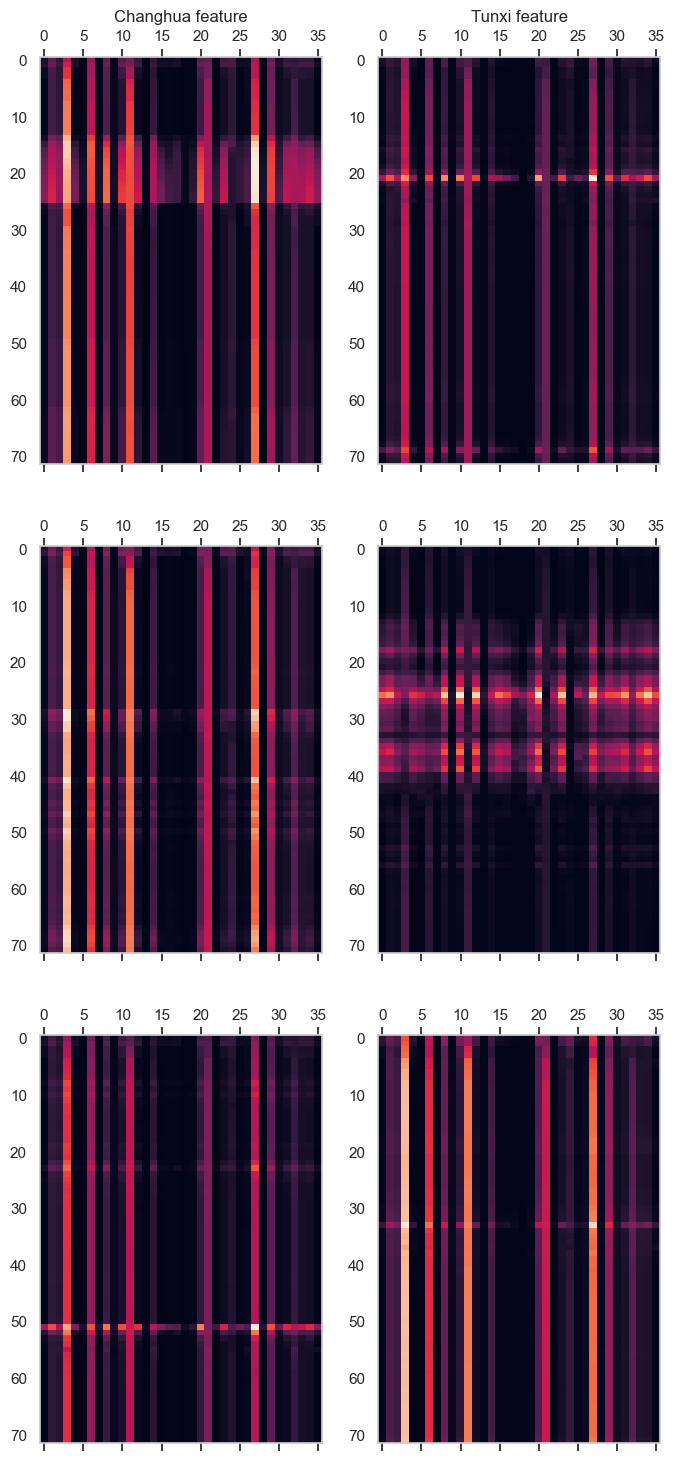}
        \caption{Comparison of target features (left) and source features (right).}
        \label{fig:feature map}
    \end{figure}
    
    \begin{figure}[h]
    \centering
        \includegraphics[width=0.4\textwidth]{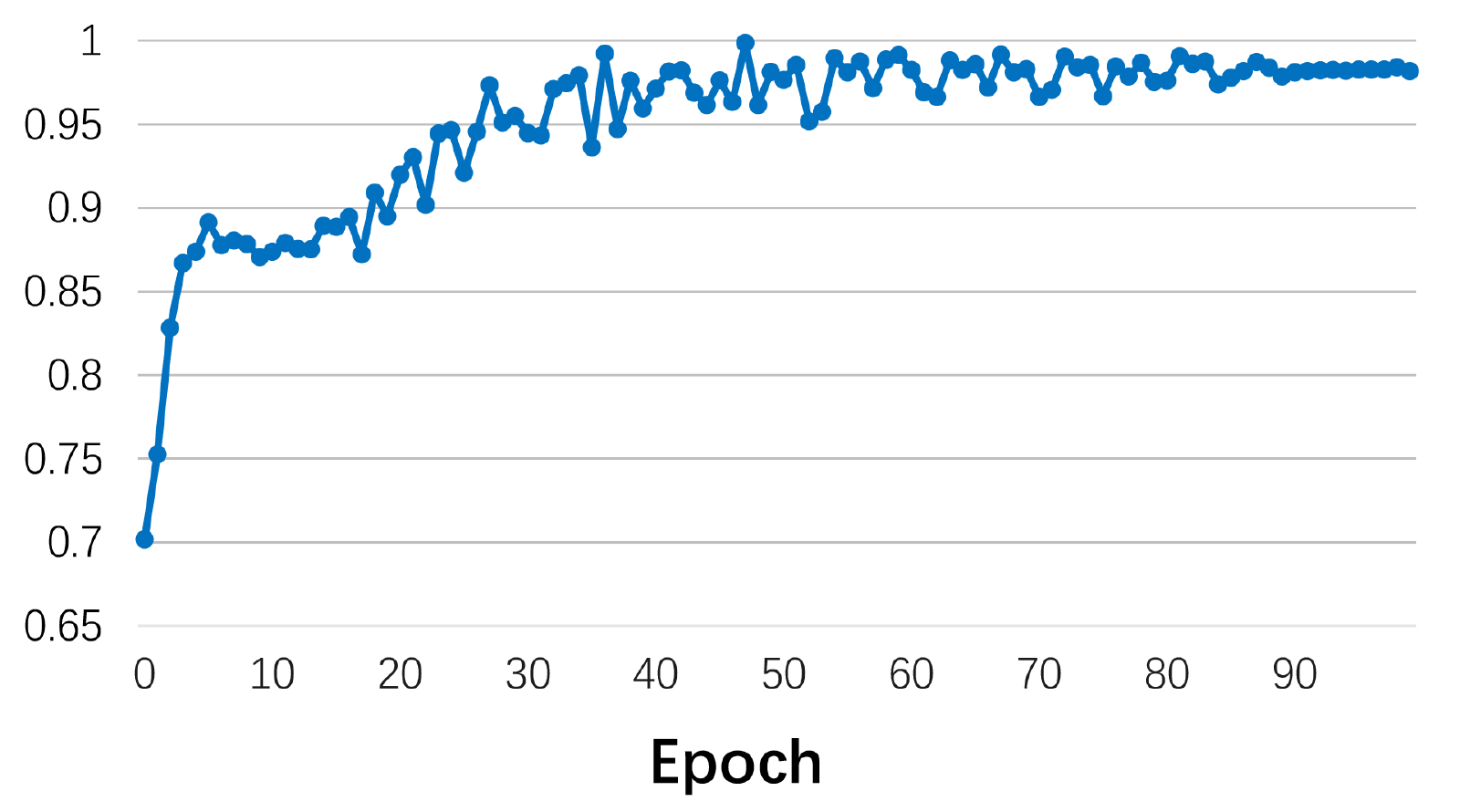}
        \caption{The curve of $\frac{{Y}_{j}}{\hat{Y}_{j}}$ during adversarial adaptation.}
        \label{fig:w}
    \end{figure}
    
    Table I shows all the experimental results. From the result of fully training set learning, the performance of the joint encoder is much lower than that of the independent encoder. However, in the supervised learning mode, whether to use residual connection has little impact on the performance of the model.
    
    We use only 2.0\% to 9.0\% of the training set sample for experiments and establishing the baseline. To reduce the impact of data quality, we have done experiments on the scale of each training set twenty times and random sampling is used to determine the training samples. Finally, we use the average of twenty results as the baseline. Fig. \ref{fig:sample_scale} presents the results of these experiments. As the amount of data decreases, the performance of the model deteriorates. This is a common problem with supervised learning.
    
    From Table I, the DC of the model trained by our approach reaches 63.49. In other words, we do not use any labels to make the model's performance equivalent to the result of supervised training with 450h to 500h samples.

\subsection{Qualitative Analysis}    
    
    Using Tunxi dataset for pretraining, we get a source encoder $E_{\text{source}}$ and a source prediction head $h_{\text{source}}$. After adversarial domain adaptation, we get a target encoder $E_{\text{target}}$. Finally we test the performance of the model composed of $E_{\text{target}}$ and $h_{\text{source}}$. We shows the results in Fig. \ref{fig:Changhua_result}. The model can accurately predict the small-scale runoff change in Changhua. And most of the trends are predicted correctly. As can be seen from the results,  $E_{\text{target}}$ can effectively extract the features of rainfall and $h_{\text{source}}$ can accurately map the features and historical runoff to the future runoff.
   
   Fig. \ref{fig:feature map} is the comparison diagram of source features and target features after adversarial domain adaptation. Their distribution is very similar, which proves that the feature output of the target encoder is already in the same feature space as the feature output of the source encoder. 
    
\subsection{Adversarial Learning Procedure}
    
    At the end of each epoch of adversarial learning, we test the model which consists of the target encoder and the source prediction head with target dataset. And we record the change in the result of $\frac{{Y}_{j}}{\hat{Y}_{j}}$, which is shown in Fig. \ref{fig:w}. As the gap between source domain and target domain becomes smaller, the performance of the model is improved. This result shows that our proposed approach achieves the desired goal.

\section{Conclusion}
    \label{sec:conclusion}
    In this technical report, we propose an unsupervised flood forecasting approach based on adversarial domain adaptation. This is the first application of unsupervised learning in flood forecasting. By pretraining and adversarial domain adaptation, our model can achieve the same level of performance as 450-500 hours of supervised learning. It is of great significance and value for flood forecasting in areas lacking hydrological data. In the future, we expect to combine unsupervised learning with few-shot to achieve a flood forecasting approach with lower data cost.
    
\bibliographystyle{IEEEtran}
\bibliography{FloodDAN}

\begin{thebibliography}{10}
\providecommand{\url}[1]{#1}
\csname url@samestyle\endcsname
\providecommand{\newblock}{\relax}
\providecommand{\bibinfo}[2]{#2}
\providecommand{\BIBentrySTDinterwordspacing}{\spaceskip=0pt\relax}
\providecommand{\BIBentryALTinterwordstretchfactor}{4}
\providecommand{\BIBentryALTinterwordspacing}{\spaceskip=\fontdimen2\font plus
\BIBentryALTinterwordstretchfactor\fontdimen3\font minus
  \fontdimen4\font\relax}
\providecommand{\BIBforeignlanguage}[2]{{%
\expandafter\ifx\csname l@#1\endcsname\relax
\typeout{** WARNING: IEEEtran.bst: No hyphenation pattern has been}%
\typeout{** loaded for the language `#1'. Using the pattern for}%
\typeout{** the default language instead.}%
\else
\language=\csname l@#1\endcsname
\fi
#2}}
\providecommand{\BIBdecl}{\relax}
\BIBdecl

\bibitem{DBLP:journals/envsoft/CostabileM15}
\BIBentryALTinterwordspacing
P.~Costabile and F.~Macchione, ``Enhancing river model set-up for 2-d dynamic
  flood modelling,'' \emph{Environ. Model. Softw.}, vol.~67, pp. 89--107, 2015.
  [Online]. Available: \url{https://doi.org/10.1016/j.envsoft.2015.01.009}
\BIBentrySTDinterwordspacing

\bibitem{Collischonn2005ForecastingRU}
W.~Collischonn, R.~Haas, I.~Andreolli, and C.~E.~M. Tucci, ``Forecasting river
  uruguay flow using rainfall forecasts from a regional weather-prediction
  model,'' \emph{Journal of Hydrology}, vol. 305, pp. 87--98, 2005.

\bibitem{Salas2018TowardsRC}
F.~R. Salas, M.~A. Somos-Valenzuela, A.~Dugger, D.~R. Maidment, D.~J. Gochis,
  C.~H. David, W.~Yu, D.~bi~Ding, E.~P. Clark, and N.~S. Noman, ``Towards
  real-time continental scale streamflow simulation in continuous and discrete
  space.'' \emph{Journal of The American Water Resources Association}, vol.~54,
  pp. 7--27, 2018.

\bibitem{Cheng2015DailyRR}
C.~tian Cheng, W.~jing Niu, F.~Zhongkai, J.~Shen, and K.~wing Chau, ``Daily
  reservoir runoff forecasting method using artificial neural network based on
  quantum-behaved particle swarm optimization,'' \emph{Water}, vol.~7, pp.
  4232--4246, 2015.

\bibitem{SHOAIB2016211}
\BIBentryALTinterwordspacing
M.~Shoaib, A.~Y. Shamseldin, B.~W. Melville, and M.~M. Khan, ``A comparison
  between wavelet based static and dynamic neural network approaches for runoff
  prediction,'' \emph{Journal of Hydrology}, vol. 535, pp. 211--225, 2016.
  [Online]. Available:
  \url{https://www.sciencedirect.com/science/article/pii/S0022169416300166}
\BIBentrySTDinterwordspacing

\bibitem{DBLP:journals/corr/abs-1908-02781}
\BIBentryALTinterwordspacing
A.~Mosavi, P.~{\"{O}}zt{\"{u}}rk, and K.~Chau, ``Flood prediction using machine
  learning models: Literature review,'' \emph{CoRR}, vol. abs/1908.02781, 2019.
  [Online]. Available: \url{http://arxiv.org/abs/1908.02781}
\BIBentrySTDinterwordspacing

\bibitem{Sulaiman18Heavy}
J.~Sulaiman and S.~Wahab, \emph{Heavy Rainfall Forecasting Model Using
  Artificial Neural Network for Flood Prone Area}, 01 2018, pp. 68--76.

\bibitem{DBLP:journals/eswa/GuoZQZL11}
\BIBentryALTinterwordspacing
J.~Guo, J.~Zhou, H.~Qin, Q.~Zou, and Q.~Li, ``Monthly streamflow forecasting
  based on improved support vector machine model,'' \emph{Expert Syst. Appl.},
  vol.~38, no.~10, pp. 13\,073--13\,081, 2011. [Online]. Available:
  \url{https://doi.org/10.1016/j.eswa.2011.04.114}
\BIBentrySTDinterwordspacing

\bibitem{SHU200831}
\BIBentryALTinterwordspacing
C.~Shu and T.~Ouarda, ``Regional flood frequency analysis at ungauged sites
  using the adaptive neuro-fuzzy inference system,'' \emph{Journal of
  Hydrology}, vol. 349, no.~1, pp. 31--43, 2008. [Online]. Available:
  \url{https://www.sciencedirect.com/science/article/pii/S0022169407006154}
\BIBentrySTDinterwordspacing

\bibitem{Nourani2014ApplicationsOH}
V.~Nourani, A.~H. Baghanam, J.~F. Adamowski, and O.~Kisi, ``Applications of
  hybrid wavelet–artificial intelligence models in hydrology: A review,''
  \emph{Journal of Hydrology}, vol. 514, pp. 358--377, 2014.

\bibitem{DBLP:journals/corr/abs-2009-14379}
\BIBentryALTinterwordspacing
T.~Iwata and A.~Kumagai, ``Few-shot learning for time-series forecasting,''
  \emph{CoRR}, vol. abs/2009.14379, 2020. [Online]. Available:
  \url{https://arxiv.org/abs/2009.14379}
\BIBentrySTDinterwordspacing

\bibitem{9515293}
D.~Chen, F.~Liu, Z.~Zhang, X.~Lu, and Z.~Li, ``Significant wave height
  prediction based on wavelet graph neural network,'' in \emph{2021 IEEE 4th
  International Conference on Big Data and Artificial Intelligence (BDAI)},
  2021, pp. 80--85.

\bibitem{7966716}
F.~Liu, F.~Xu, and S.~Yang, ``A flood forecasting model based on deep learning
  algorithm via integrating stacked autoencoders with bp neural network,'' in
  \emph{2017 IEEE Third International Conference on Multimedia Big Data
  (BigMM)}, 2017, pp. 58--61.

\bibitem{Dineva14Fuzzy}
A.~Dineva, A.~R. Várkonyi-Kóczy, and J.~K. Tar, ``Fuzzy expert system for
  automatic wavelet shrinkage procedure selection for noise suppression,'' pp.
  163--168, 2014.

\bibitem{pan2009survey}
S.~J. Pan and Q.~Yang, ``A survey on transfer learning,'' \emph{IEEE
  Transactions on knowledge and data engineering}, vol.~22, no.~10, pp.
  1345--1359, 2009.

\bibitem{long2015learning}
M.~Long, Y.~Cao, J.~Wang, and M.~Jordan, ``Learning transferable features with
  deep adaptation networks,'' in \emph{International conference on machine
  learning}.\hskip 1em plus 0.5em minus 0.4em\relax PMLR, 2015, pp. 97--105.

\bibitem{long2016unsupervised}
M.~Long, H.~Zhu, J.~Wang, and M.~I. Jordan, ``Unsupervised domain adaptation
  with residual transfer networks,'' \emph{arXiv preprint arXiv:1602.04433},
  2016.

\bibitem{DBLP:journals/corr/ChenLCHFS17}
\BIBentryALTinterwordspacing
T.~Chen, Y.~Liao, C.~Chuang, W.~T. Hsu, J.~Fu, and M.~Sun, ``Show, adapt and
  tell: Adversarial training of cross-domain image captioner,'' \emph{CoRR},
  vol. abs/1705.00930, 2017. [Online]. Available:
  \url{http://arxiv.org/abs/1705.00930}
\BIBentrySTDinterwordspacing

\bibitem{DBLP:journals/corr/abs-2005-00791}
\BIBentryALTinterwordspacing
D.~Ghosal, D.~Hazarika, A.~Roy, N.~Majumder, R.~Mihalcea, and S.~Poria,
  ``Kingdom: Knowledge-guided domain adaptation for sentiment analysis,''
  \emph{CoRR}, vol. abs/2005.00791, 2020. [Online]. Available:
  \url{https://arxiv.org/abs/2005.00791}
\BIBentrySTDinterwordspacing

\bibitem{DBLP:conf/icml/GaninL15}
\BIBentryALTinterwordspacing
Y.~Ganin and V.~S. Lempitsky, ``Unsupervised domain adaptation by
  backpropagation,'' in \emph{Proceedings of the 32nd International Conference
  on Machine Learning, {ICML} 2015, Lille, France, 6-11 July 2015}, ser. {JMLR}
  Workshop and Conference Proceedings, F.~R. Bach and D.~M. Blei, Eds.,
  vol.~37.\hskip 1em plus 0.5em minus 0.4em\relax JMLR.org, 2015, pp.
  1180--1189. [Online]. Available:
  \url{http://proceedings.mlr.press/v37/ganin15.html}
\BIBentrySTDinterwordspacing

\bibitem{DBLP:journals/jmlr/GaninUAGLLML16}
\BIBentryALTinterwordspacing
Y.~Ganin, E.~Ustinova, H.~Ajakan, P.~Germain, H.~Larochelle, F.~Laviolette,
  M.~Marchand, and V.~S. Lempitsky, ``Domain-adversarial training of neural
  networks,'' \emph{J. Mach. Learn. Res.}, vol.~17, pp. 59:1--59:35, 2016.
  [Online]. Available: \url{http://jmlr.org/papers/v17/15-239.html}
\BIBentrySTDinterwordspacing

\bibitem{DBLP:journals/ijon/WangD18}
\BIBentryALTinterwordspacing
M.~Wang and W.~Deng, ``Deep visual domain adaptation: {A} survey,''
  \emph{Neurocomputing}, vol. 312, pp. 135--153, 2018. [Online]. Available:
  \url{https://doi.org/10.1016/j.neucom.2018.05.083}
\BIBentrySTDinterwordspacing

\bibitem{DBLP:journals/corr/abs-2010-03978}
\BIBentryALTinterwordspacing
A.~Farahani, S.~Voghoei, K.~Rasheed, and H.~R. Arabnia, ``A brief review of
  domain adaptation,'' \emph{CoRR}, vol. abs/2010.03978, 2020. [Online].
  Available: \url{https://arxiv.org/abs/2010.03978}
\BIBentrySTDinterwordspacing

\bibitem{DBLP:conf/coling/RamponiP20}
\BIBentryALTinterwordspacing
A.~Ramponi and B.~Plank, ``Neural unsupervised domain adaptation in {NLP} - {A}
  survey,'' in \emph{Proceedings of the 28th International Conference on
  Computational Linguistics, {COLING} 2020, Barcelona, Spain (Online), December
  8-13, 2020}, D.~Scott, N.~Bel, and C.~Zong, Eds.\hskip 1em plus 0.5em minus
  0.4em\relax International Committee on Computational Linguistics, 2020, pp.
  6838--6855. [Online]. Available:
  \url{https://doi.org/10.18653/v1/2020.coling-main.603}
\BIBentrySTDinterwordspacing

\bibitem{DBLP:journals/corr/abs-2102-09508}
\BIBentryALTinterwordspacing
H.~Guan and M.~Liu, ``Domain adaptation for medical image analysis: {A}
  survey,'' \emph{CoRR}, vol. abs/2102.09508, 2021. [Online]. Available:
  \url{https://arxiv.org/abs/2102.09508}
\BIBentrySTDinterwordspacing

\bibitem{Gulrajani2017NIPS_improved}
I.~Gulrajani, F.~Ahmed, M.~Arjovsky, V.~Dumoulin, and A.~C. Courville,
  ``Improved training of wasserstein gans,'' in \emph{Advances in Neural
  Information Processing Systems 30: Annual Conference on Neural Information
  Processing Systems 2017, December 4-9, 2017, Long Beach, CA, {USA}},
  I.~Guyon, U.~von Luxburg, S.~Bengio, H.~M. Wallach, R.~Fergus, S.~V.~N.
  Vishwanathan, and R.~Garnett, Eds., 2017, pp. 5767--5777.

\bibitem{9451544}
Z.~Li, F.~Liu, W.~Yang, S.~Peng, and J.~Zhou, ``A survey of convolutional
  neural networks: Analysis, applications, and prospects,'' \emph{IEEE
  Transactions on Neural Networks and Learning Systems}, pp. 1--21, 2021.

\bibitem{jcst2022self}
F.~Liu, D.~Chen, R.~Zhou, S.~Yanh, and F.~Xu, ``Self-supervised music motion
  synchronization learning for music-driven conducting motion generation,''
  \emph{Journal of Computer Science and Technology}, 2022.

\end{thebibliography}

\end{document}